\documentclass{article}

\usepackage{PRIMEarxiv}

\usepackage[utf8]{inputenc} 
\usepackage[T1]{fontenc}    
\usepackage{hyperref}       
\usepackage{url}            
\usepackage{booktabs}       
\usepackage{amsfonts}       
\usepackage{nicefrac}       
\usepackage{microtype}      
\usepackage{lipsum}
\usepackage{fancyhdr}       
\usepackage{graphicx}       
\graphicspath{{media/}}     

\usepackage{cite}
\usepackage{amsmath,amssymb,amsfonts}
\usepackage{algorithm}
\usepackage{algpseudocode}
\usepackage{graphicx}
\usepackage{textcomp}
\usepackage{xcolor}
\usepackage{tikz}
\usepackage{amsmath}
\usepackage{graphicx}
\usetikzlibrary{arrows.meta, positioning, shapes.geometric, calc, decorations.pathreplacing}
\def\BibTeX{{\rm B\kern-.05em{\sc i\kern-.025em b}\kern-.08em
    T\kern-.1667em\lower.7ex\hbox{E}\kern-.125emX}}

\title{SPIRAL: Self-Play Incremental Racing Algorithm for Learning in Multi-Drone Competitions
\thanks{
\textbf{Accepted for publication at the 2025 IEEE Innovations in Intelligent Systems and Applications (ASYU) Conference.}} 
}

\author{
  Onur Akgün \\
  Department of Mechatronics Engineering \\
  Faculty of Engineering \\
  Turkish-German University \\
  Istanbul, Türkiye\\
  \texttt{akgun@tau.edu.tr} \\
}

\begin{document}
\maketitle

\begin{abstract}
This paper introduces SPIRAL (Self-Play Incremental Racing Algorithm for Learning), a novel approach for training autonomous drones in multi-agent racing competitions. SPIRAL distinctively employs a self-play mechanism to incrementally cultivate complex racing behaviors within a challenging, dynamic environment. Through this self-play core, drones continuously compete against increasingly proficient versions of themselves, naturally escalating the difficulty of competitive interactions. This progressive learning journey guides agents from mastering fundamental flight control to executing sophisticated cooperative multi-drone racing strategies. Our method is designed for versatility, allowing integration with any state-of-the-art Deep Reinforcement Learning (DRL) algorithms within its self-play framework. Simulations demonstrate the significant advantages of SPIRAL and benchmark the performance of various DRL algorithms operating within it. Consequently, we contribute a versatile, scalable, and self-improving learning framework to the field of autonomous drone racing. SPIRAL's capacity to autonomously generate appropriate and escalating challenges through its self-play dynamic offers a promising direction for developing robust and adaptive racing strategies in multi-agent environments. This research opens new avenues for enhancing the performance and reliability of autonomous racing drones in increasingly complex and competitive scenarios.
\end{abstract}


\section{Introduction}
Autonomous drone racing has rapidly gained prominence as a compelling research frontier, challenging the boundaries of autonomous flight in intricate and rapidly changing environments \cite{moon2019challenges, foehn2022alphapilot}. The advent of the deep learning era has catalyzed significant progress in single-drone racing, particularly in enhancing perception and flight dynamics \cite{kaufmann2018deep, jung2018perception}. These advancements have culminated in successful demonstrations of end-to-end learning for autonomous vision-based drone racing \cite{loquercio2019deep}.

For the effective end-to-end learning and reliable sim-to-real transfer of policies for vision-based autonomous drone racing, the use of photorealistic simulation frameworks is indispensable, as online learning is often unfeasible due to the high speeds and inherent risk of collisions. Prominent open-source simulation platforms such as FlightGoggles \cite{guerra2019flightgoggles}, Flightmare \cite{song2021flightmare}, and Agilicious \cite{foehn2022agilicious} provide the necessary environments for such development. Leveraging these sophisticated simulation tools, end-to-end learning methodologies have enabled deep reinforcement learning agents to achieve champion-level performance in drone racing, illustrating the potential for learning-based control to surpass expert human pilots \cite{kaufmann2023champion}. Furthermore, comprehensive analyses comparing optimal control and reinforcement learning techniques in autonomous racing have continued to elevate the benchmarks of achievable performance \cite{song2023reaching}.

\begin{figure}[t!]
    \centering
    \includegraphics[width=\linewidth]{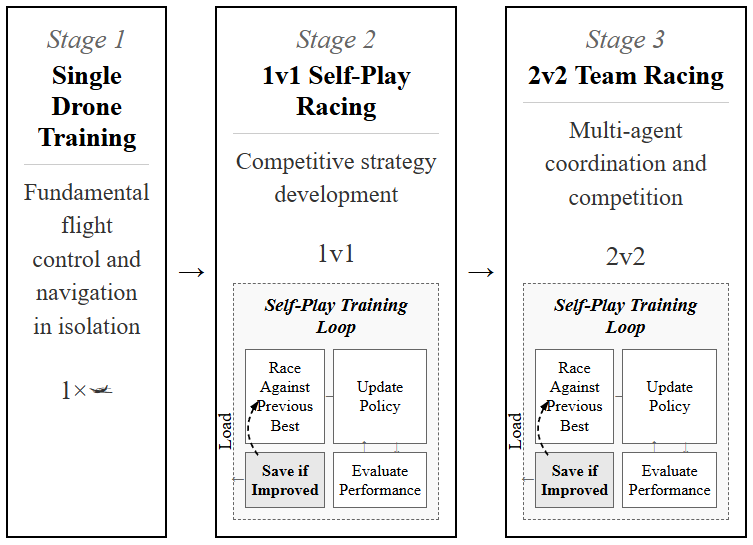}
    \caption{The SPIRAL (Self-Play Incremental Racing Algorithm for Learning) training framework consists of three progressive stages. Stage 1 establishes fundamental flight control capabilities in isolation. Stages 2 and 3 employ self-play training loops where agents continuously improve by racing against their previous best policies, with Stage 2 focusing on 1v1 competition and Stage 3 extending to 2v2 team dynamics. The self-play mechanism (shown in dashed boxes) enables continuous improvement without requiring external expert demonstrations.}
    \label{fig:spiral_diagram}
\end{figure}

Transitioning from single-agent scenarios, multi-drone racing introduces a new echelon of challenges. These include the complexities of inter-agent coordination, robust collision avoidance in cluttered spaces, and the development of sophisticated competitive strategies, which significantly extend beyond those encountered in general multi-agent problems like flocking or basic swarm behaviors \cite{huttenrauch2019deep, vasarhelyi2018optimized}. While recent contributions have explored game-theoretic planners \cite{spica2020real, wang2020multi, di2023cooperative}, which demonstrate state-of-the-art performance, they often incur substantial computational costs. This highlights the demand for methods that are not only competitive but also data-efficient and scalable, capable of transitioning seamlessly from single to multi-agent racing and from simpler to more complex environmental settings. To address this, we advocate for self-play learning as a transformative training paradigm for multi-agent drone racing.

The research presented in this paper draws inspiration from the remarkable successes of self-play mechanisms in complex competitive domains, such as the games of Chess and Go \cite{silver2017mastering}. This paradigm has since been extended to achieve superhuman performance in highly complex multi-agent environments such as Dota 2 \cite{berner2019dota} and StarCraft II \cite{vinyals2019grandmaster}, demonstrating its power and scalability. A pivotal development in this area was Neural Fictitious Self-Play (NFSP), introduced by Heinrich and Silver \cite{heinrich2016deep}, which offered a scalable, end-to-end learning approach to approximate Nash equilibria without requiring prior domain-specific knowledge. NFSP demonstrated its efficacy in intricate imperfect-information games like poker by synergizing fictitious self-play with deep reinforcement learning. Grounded in the advancements of vision-based autonomous drone racing, the availability of high-fidelity simulation platforms, and the foundational principles of NFSP, this work introduces SPIRAL (Self-Play Incremental Racing Algorithm for Learning). SPIRAL is a novel self-play based, incremental multi-agent learning framework tailored for multi-drone racing environments. It is designed to be scalable and, as indicated by initial numerical experiments, capable of achieving competitive performance. Our framework allows for the integration of various state-of-the-art Deep Reinforcement Learning (DRL) algorithms, such as those available in established libraries like Stable-Baselines3 \cite{stable-baselines3} to continually refine drone capabilities.

\section{Related Works}
\label{sec:regress:RW}

Recent progress in multi-drone racing has focused on game-theoretic planning. Spica et al. \cite{spica2020real} introduced a real-time planner for two-player scenarios that approximates a Nash equilibrium using a sensitivity term to predict opponent actions, outperforming model predictive control. Building on this, Wang et al. \cite{wang2020multi} extended the approach to N-player races in 3D, using an iterative best response algorithm to find a collective Nash equilibrium. They managed the computational complexity using parallelization and a neural network for trajectory prediction.

Addressing cooperative strategies, Di et al. \cite{di2023cooperative} proposed a leader-wingman framework. By combining a game-theoretic planner for the leader with a line-of-sight cooperative method for the wingman, their approach increases the team's win probability against faster individual rivals while maintaining lower computational overhead than purely game-theoretic strategies.

Collectively, these state-of-the-art works show a trend towards sophisticated game-theoretic frameworks that model competitive and cooperative interactions. However, they highlight persistent challenges in balancing computational feasibility with strategic depth and ensuring adaptability in dynamic environments. This landscape motivates our exploration of self-play as an alternative learning paradigm to foster robust and adaptive strategies more efficiently.

\section{Methodology}
\label{sec:regress:methodology}

This section details the proposed Self-Play Incremental Racing Algorithm for Learning (SPIRAL), including the underlying problem formulation and the specifics of the learning framework.

\subsection{Problem Formulation: Multi-Drone Racing as a Markov Decision Process}

We formulate the multi-drone racing problem as a decentralized multi-agent Markov Decision Process (MDP). For each individual agent (drone) $i$, the problem can be described by a tuple $(S_i, A_i, P_i, R_i, \gamma)$, where:

\begin{itemize}
    \item $S_i$ is the observation space for agent $i$. This space encompasses information about the agent's own state, the state of nearby opponents, and relevant aspects of the race environment (e.g., gate locations).
    \item $A_i$ is the action space for agent $i$, defining the control inputs the drone can execute.
    \item $P_i(s'{i} | s_i, \mathbf{a})$ is the state transition probability function, denoting the probability of agent $i$ transitioning to state $s'{i}$ given its current state $s_i$ and the joint action $\mathbf{a} = (a_1, ..., a_N)$ of all $N$ agents in the environment. 
    \item $R_i(s_i, a_i, s'{i})$ is the reward function for agent $i$, which quantifies the immediate feedback received after taking action $a_i$ in state $s_i$ and transitioning to state $s'{i}$. This function is designed to incentivize progress through the race course, penalize undesirable events (e.g., collisions), and potentially reward strategic behaviors like overtaking opponents.
    \item $\gamma \in$ is the discount factor, which balances the importance of immediate versus future rewards.
\end{itemize}

Each agent $i$ aims to learn a policy $\pi_i: S_i \rightarrow A_i$ (or $\pi_i: S_i \rightarrow \mathcal{P}(A_i)$ for stochastic policies, where $\mathcal{P}(A_i)$ is the set of probability distributions over $A_i$) that maximizes its expected discounted cumulative reward, $J_i(\pi_i) = \mathbb{E}{\pi_i, \pi{-i}, P} \left[ \sum_{t=0}^{T} \gamma^t R_i(s_{i,t}, a_{i,t}, s_{i,t+1}) \right]$, where $\pi_{-i}$ represents the policies of all other agents.

\subsection{Simulation Environment and Drone Control Model}
\label{subsec:simulation_environment}

All experiments are conducted within the PyBullet physics simulator, which provides a high-fidelity and computationally efficient environment for modeling multi-body dynamics. The core of our simulation is a standard quadrotor model, whose motion is governed by well-established rigid-body dynamics.

To describe the drone's motion, we define two essential coordinate frames: an inertial world frame, $W = {O_W, \mathbf{x}_W, \mathbf{y}_W, \mathbf{z}_W}$, and a body frame, $B = {O_B, \mathbf{x}_B, \mathbf{y}_B, \mathbf{z}_B}$, fixed to the drone's center of mass. The state of the quadrotor at any time $t$ is fully described by its position, velocity, orientation, and angular velocity. This can be represented by the state vector $\mathbf{x} = [\mathbf{p}^T, \dot{\mathbf{p}}^T, \boldsymbol{\Phi}^T, \boldsymbol{\omega}^T]^T \in \mathbb{R}^{12}$, where $\mathbf{p} = [x, y, z]^T$ is the position and $\dot{\mathbf{p}} = [v_x, v_y, v_z]^T$ is the linear velocity in the world frame $W$. The orientation is given by the Euler angles $\boldsymbol{\Phi} = [\phi, \theta, \psi]^T$ (roll, pitch, yaw), and $\boldsymbol{\omega} = [p, q, r]^T$ is the angular velocity in the body frame $B$. The orientation is also represented by the rotation matrix $\mathbf{R} \in SO(3)$ which transforms vectors from the body frame to the world frame.

The translational dynamics are governed by Newton's second law, describing the acceleration of the drone's center of mass:

\begin{equation}
    m\ddot{\mathbf{p}} = \begin{bmatrix} 0 \ 0 \ -mg \end{bmatrix} + \mathbf{R}\mathbf{T}_B - \mathbf{K}_d \dot{\mathbf{p}}
\end{equation}

In this equation, $m$ represents the total mass of the drone and $g$ is the gravitational acceleration. The term $\mathbf{R}\mathbf{T}_B$ is the total thrust vector, generated by the four rotors, projected into the world frame. The thrust $\mathbf{T}_B = [0, 0, T]^T$ acts along the drone's body z-axis, where $T$ is the sum of the forces produced by each of the four motors. The final term, $\mathbf{K}_d \dot{\mathbf{p}}$, models the effects of aerodynamic drag as a force opposing the drone's linear velocity, with $\mathbf{K}_d$ being a diagonal matrix of drag coefficients.

The rotational dynamics are described by the Newton-Euler equation for rigid bodies:

\begin{equation}
    \mathbf{I}\dot{\boldsymbol{\omega}} = \boldsymbol{\tau} - \boldsymbol{\omega} \times (\mathbf{I}\boldsymbol{\omega})
\end{equation}

Here, $\mathbf{I} \in \mathbb{R}^{3 \times 3}$ is the symmetric, positive-definite inertia matrix of the drone. The term $\boldsymbol{\tau} = [\tau_\phi, \tau_\theta, \tau_\psi]^T$ represents the vector of control torques around the body frame axes, which are generated by creating differential thrust among the four rotors. The term $\boldsymbol{\omega} \times (\mathbf{I}\boldsymbol{\omega})$ accounts for the gyroscopic effects due to the rotation of the drone's body.

The collective thrust $T$ and control torques $\boldsymbol{\tau}$ are produced by the individual forces $f_j$ of the four motors. This relationship is defined by the control allocation matrix, which maps motor forces to generalized forces:

\begin{align}
    \begin{bmatrix}
    T \\
    \boldsymbol{\tau}
\end{bmatrix}
=
\begin{bmatrix}
    k_f & k_f & k_f & k_f \\
    0 & l k_f & 0 & -l k_f \\
    -l k_f & 0 & l k_f & 0 \\
    k_m & -k_m & k_m & -k_m
\end{bmatrix}
\begin{bmatrix}
    \omega_1^2 \\
    \omega_2^2 \\
    \omega_3^2 \\
    \omega_4^2
\end{bmatrix}
\end{align}

where $\omega_j$ is the angular velocity of motor $j$, $l$ is the distance from the center of mass to each motor, $k_f$ is the thrust coefficient, and $k_m$ is the drag torque coefficient.

Given this dynamic model, we employ a hierarchical control architecture to separate high-level strategic decision-making from low-level flight stabilization. The reinforcement learning agent constitutes the high-level policy, operating at a frequency of approximately 50 Hz. As detailed in Section \ref{subsubsec:action_space}, this policy outputs a desired state reference $a_{i,t} = [p^{x,t}, p^{y,t}, p^{z,t}, \psi^{t}]^T$. This command is then passed to a high-frequency (e.g., 240 Hz) low-level PID controller. This embedded controller continuously calculates the error between the desired state $(p^, \psi^)$ and the current measured state from the simulator. Based on this error, it computes the necessary total thrust $T$ and control torques $\boldsymbol{\tau}$ and allocates the required RPMs to the individual motors to achieve them. This architecture allows the RL agent to focus on the strategic task of selecting optimal waypoints, while the PID controller robustly handles the complex, fast-paced dynamics of flight stabilization.

\subsection{SPIRAL: Self-Play Incremental Racing Algorithm for Learning}

We introduce SPIRAL (Self-Play Incremental Racing Algorithm for Learning), a training framework designed to efficiently develop sophisticated racing capabilities in autonomous drones. SPIRAL integrates two key concepts: self-play and an incremental increase in scenario complexity. The self-play mechanism allows agents to learn by competing against evolving versions of themselves. The incremental complexity is achieved by gradually making the racing scenarios more challenging, guiding the learning process from fundamental control to advanced multi-agent maneuvers.

The core idea of SPIRAL is to create a learning environment where an agent, continually improves by facing opponents that are themselves improving or represent diverse, challenging strategies. This iterative refinement process naturally leads to the emergence of complex behaviors.

\subsubsection{State Space Representation ($S_i$)}
\label{subsubsec:state_space}

The observation vector $s_{i,t}$ for agent $i$ provides sufficient information for decision-making by combining two primary components. The first is the drone's 12D ego-state, which describes its physical status:
$s_{ego,t} = [p_{x,t}, p_{y,t}, p_{z,t}, v_{x,t}, v_{y,t}, v_{z,t}, \phi_t, \theta_t, \psi_t, \omega_{x,t}, \omega_{y,t}, \omega_{z,t}]^T \in \mathbb{R}^{12}$
This vector includes the drone's Cartesian position, linear velocity, orientation (roll, pitch, yaw), and angular velocity.

The second component is a contextual information buffer, $B_{i,t}$, which provides temporal awareness by storing a history of key features from the past $K$ time steps. This buffer includes relative position data for the $M_o$ closest opponents and $M_g$ upcoming gates, as well as the agent's own recent actions. The full observation is a concatenation of the current ego-state and processed information from this buffer: $s_{i,t} = \text{concat}(s_{ego,t}, f(B_{i,t}))$.

In our implementation, we set $M_o=2$ and $M_g=2$. The buffer stores these features for the past $K=50$ time steps, a value chosen empirically to balance the need for sufficient historical context against the risk of excessive input dimensionality.

\subsubsection{Action Space ($A_i$)}
\label{subsubsec:action_space}

The action $a_{i,t}$ selected by the agent is a continuous 4D vector consisting of high-level control targets:
$a_{i,t} = [p^{x,t}, p^{y,t}, p^{z,t}, \psi^{t}]^T \in \mathbb{R}^4$
where $(p^{x,t}, p^{y,t}, p^{z,t})$ is the desired target 3D position and $\psi^{t}$ is the desired target yaw angle.

This hierarchical control structure abstracts the complexities of low-level flight dynamics. The high-level commands are translated by a low-level PID controller into the necessary motor inputs, allowing the DRL agent to focus on strategic decision-making rather than flight stabilization.

\subsubsection{Reward Function ($R_i$)}
\label{subsubsec:reward_function}
The reward function is a composite function designed to guide the learning agent towards desired racing behaviors by penalizing detrimental actions. The total reward is a weighted sum of four distinct terms:

$R_{i,t} = w_p R_{p,t} + w_c R_{c,t} + w_a R_{a,t} + R_{t,\mathrm{episode\_end}}$

where $w_p, w_c, w_a$ are non-negative weighting factors. The components are defined as follows:
\begin{enumerate}
    \item Progress through the Course ($R_{p,t}$): This component incentivizes making progress along the track and navigating through gates. It is calculated as $R_{p,t} = \alpha \cdot \Delta d_t + \beta \cdot G_t$, where $\Delta d_t$ is the change in distance along the track centerline and $G_t$ is an indicator for successfully passing a gate.
    \item Collision Avoidance ($R_{c,t}$): This component imposes a significant penalty for any collisions with obstacles, other drones, or course boundaries. It is defined as:
    $R_{c,t} =
\begin{cases}
    -C & \text{if a collision occurs} \\
    0  & \text{otherwise}
\end{cases}$
where $C$ is a large positive constant.
    \item Angular Alignment ($R_{a,t}$): This reward encourages the drone to orient itself correctly towards the next gate to facilitate smoother navigation. It is given by $R_{a,t} = \zeta \cdot \max(0, \cos(\theta_t) - \cos(\theta_{\text{threshold}}))$, where the reward is positive only if the alignment angle $\theta_t$ is less than a predefined threshold $\theta_{\text{threshold}}$.
    \item Time Reward ($R_{t, \text{episode\_end}}$): This is an episodic reward designed to encourage faster lap times, awarded only at the end of a successfully completed lap. It is calculated as $R_{t, \text{episode\_end}} = \dfrac{100}{T_{\text{lap}}}$, where $T_{\text{lap}}$ is the total time taken to complete the lap.
\end{enumerate}

The weights and scaling factors for these components are crucial hyperparameters that are tuned empirically to achieve the desired balance between speed, safety, and efficiency.

\subsubsection{Staged Training Protocol and Self-Play Dynamics}
\label{subsubsec:staged_training}

The SPIRAL framework employs a structured, multi-stage training protocol, illustrated in Figure \ref{fig:spiral_diagram}, to progressively develop drone racing capabilities through self-play and incrementally increasing task complexity. This protocol advances through three core stages:

\begin{enumerate}
    \item Single Drone Training: The agent first trains alone to master fundamental flight control, gate navigation, and solo track completion without direct opponents.
    \item 1v1 Head-to-Head Racing: The agent develops competitive strategies, such as overtaking and reacting to a dynamic opponent, by competing against its own best-performing past policies. This introduces direct competition and adaptation.
    \item 2v2 Team-Based Racing: The agent cultivates more complex behaviors by facing two opponents drawn from the self-play pool, requiring a balance of competitive tactics and implicit coordination. However, it is crucial to note that because each agent optimizes its own decentralized reward, this stage primarily tests adaptation to a crowded environment. Any emergent teamwork is a byproduct of individual, competitive goals rather than a result of true, cooperative strategies, which remains a limitation of this reward structure.
\end{enumerate}

Within each stage, the training loop involves generating experience by racing against selected self-play opponents. The agent's policy is then refined using Proximal Policy Optimization (PPO) with the collected experience. Performance is periodically assessed (e.g., by win rate and lap time), and if an improvement is observed, the current model is saved as the new best. This newly saved model is then imported as an opponent model, ensuring the agent continuously faces increasingly competent versions of itself.

This staged progression forms a natural curriculum where skills acquired in simpler stages are foundational for more complex ones. The combination of an explicit, staged curriculum with the implicit challenge escalation from self-play allows SPIRAL to efficiently guide agents towards sophisticated racing strategies.

\section{Results and Discussion}
\label{sec:results_discussion}

\subsection{Experimental Setup}
\begin{figure}[htbp]
\centering
\includegraphics[width=0.5\textwidth]{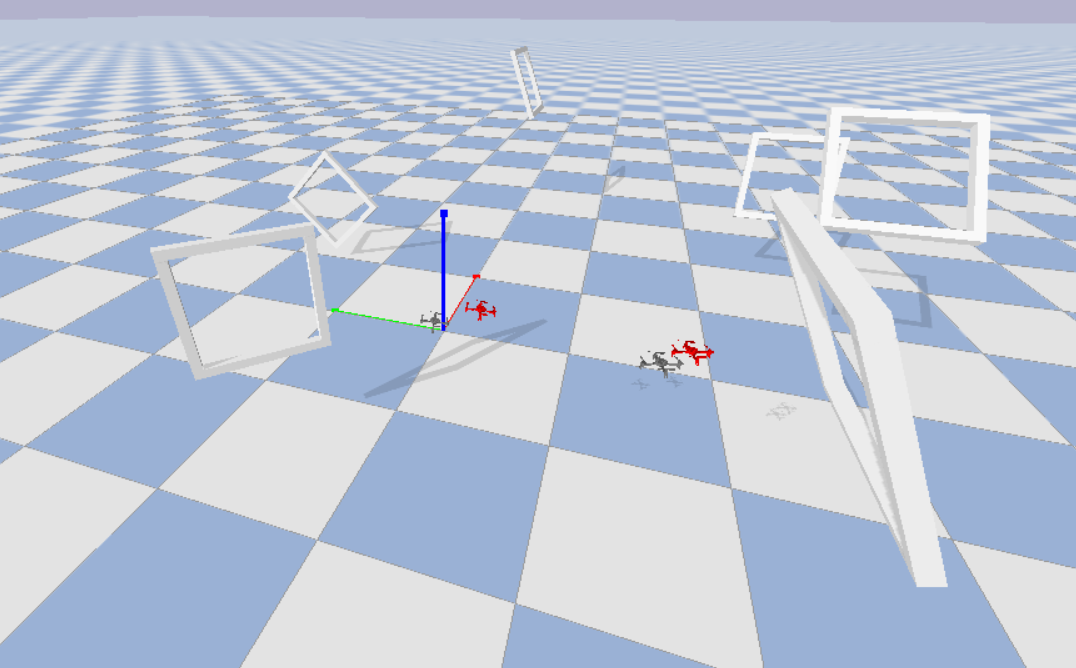}
\caption{Visualization of the 2v2 drone racing environment. The race track consists of six gates arranged in a circuit on a checkered floor. Four drones (two red and two white) are positioned on the track, representing the 2v2 team formation used in our experiments. The blue axis indicator provides spatial reference.}
\label{fig:racetrack}
\end{figure}

To empirically validate the efficacy of the proposed SPIRAL framework, a series of experiments were conducted within a simulated drone racing environment. The environment features a circuit-style racetrack with six gates, as depicted in Figure~\ref{fig:racetrack}.

Our evaluation compares SPIRAL against two key baselines: a state-of-the-art game-theoretic planner and a standard PPO agent trained without self-play. For the first baseline, we implemented the game-theoretic trajectory planner from Wang et al. \cite{wang2020multi}. As no public code was available, our implementation is based on the core mechanisms described in their work, specifically its iterative best response mechanism and opponent trajectory inference for 3D racing. The second baseline, PPO No Selfplay, employs the same network architecture and reward function as SPIRAL but is trained without the self-play mechanism. This allows us to isolate and quantify the direct benefits of the self-play dynamic.

Our comparative analysis evaluates three distinct training methodologies in a two-drone racing scenario:
\begin{itemize}
    \item PPO with Self-Play (SPIRAL): The proposed method
    \item PPO without Self-Play: A standard reinforcement learning approach without the adaptive opponent mechanism
    \item SE-IBR (Baseline): The Sensitivity-Enhanced Iterative Best Response method from Wang et al. \cite{wang2020multi}, serving as a state-of-the-art game-theoretic baseline
\end{itemize}

For each methodology, we executed 50 independent evaluation runs, each with randomized initial drone placements to ensure statistical robustness and mitigate positional bias. This rigorous experimental design facilitates a direct and fair comparison of the performance characteristics of each approach under challenging and consistent racing conditions.

\subsection{Performance Metrics}

The performance of each training methodology was assessed using two primary metrics:
\begin{itemize}
    \item Lap Time (s): The average time, in seconds, required for a drone to complete one full lap of the race course. This metric serves as a direct measure of the speed and efficiency of the learned control policy. Lower values indicate superior performance.
    \item Success Ratio: The proportion of evaluation runs in which a drone successfully completes the entire course without any collisions and correctly navigates through all gates. This metric quantifies the reliability and safety of the learned racing behaviors. Higher values indicate superior performance.
\end{itemize}

Together, these metrics provide a comprehensive evaluation, capturing the critical trade-off between speed and consistency in autonomous drone racing.

\subsection{Experimental Results}

To validate our approach, we evaluate the final policies produced by each training method in two distinct experimental setups. These setups correspond to the final two stages of our proposed training protocol: a two-drone (1v1) race and a more complex four-drone (2v2) race. The results, averaged over 50 evaluation runs with random initial placements for each scenario, are summarized in Tables ~\ref{tab:approach_comparison_2_agents_test1} and ~\ref{tab:approach_comparison_4_agents_test2}.

Table ~\ref{tab:approach_comparison_2_agents_test1} presents the performance in the two-drone (1v1) head-to-head racing scenario. This setup evaluates the core competitive abilities developed in Stage 2 of our training protocol.

\begin{table}[htbp]
\centering
\begin{tabular}{|l|c|c|}
\hline
\textbf{Approach} & \textbf{Lap Time (s)} & \textbf{Success Ratio} \\
\hline
PPO Selfplay (SPIRAL) & $\mathbf{13.7124 \pm 0.0005}$ & $0.6900 \pm 0.0500$ \\
SE-IBR (Baseline) & $14.2506 \pm 0.2986$ & $\mathbf{0.8100 \pm 0.1100}$ \\
PPO No Selfplay & $16.3134 \pm 0.0727$ & $0.8000 \pm 0.0200$ \\
\hline
\end{tabular}
\caption{Performance comparison in the Two-Drone (1v1) Racing Scenario over 50 flights with random initial placements.}
\label{tab:approach_comparison_2_agents_test1}
\end{table}

Table ~\ref{tab:approach_comparison_4_agents_test2} evaluates the agents in the more challenging four-drone (2v2) team-based racing scenario, which corresponds to Stage 3 of our protocol. This experiment is designed to test the scalability of the learned policies and their performance under increased agent density and complexity.

\begin{table}[htbp]
\centering
\begin{tabular}{|l|c|c|}
\hline
\textbf{Approach} & \textbf{Lap Time (s)} & \textbf{Success Ratio} \\
\hline
PPO Selfplay (SPIRAL) & $\mathbf{13.71 \pm 0.01}$ & $0.56 \pm 0.01$ \\
SE-IBR (Baseline) & $14.39 \pm 0.33$ & $\mathbf{0.62 \pm 0.22}$ \\
PPO No Selfplay & $17.82 \pm 0.27$ & $0.31 \pm 0.12$ \\
\hline
\end{tabular}
\caption{Performance comparison in the Four-Drone (2v2) Racing Scenario over 50 flights with random initial placements.}
\label{tab:approach_comparison_4_agents_test2}
\end{table}

\begin{figure}[htbp]
    \centering
    \includegraphics[width=\columnwidth]{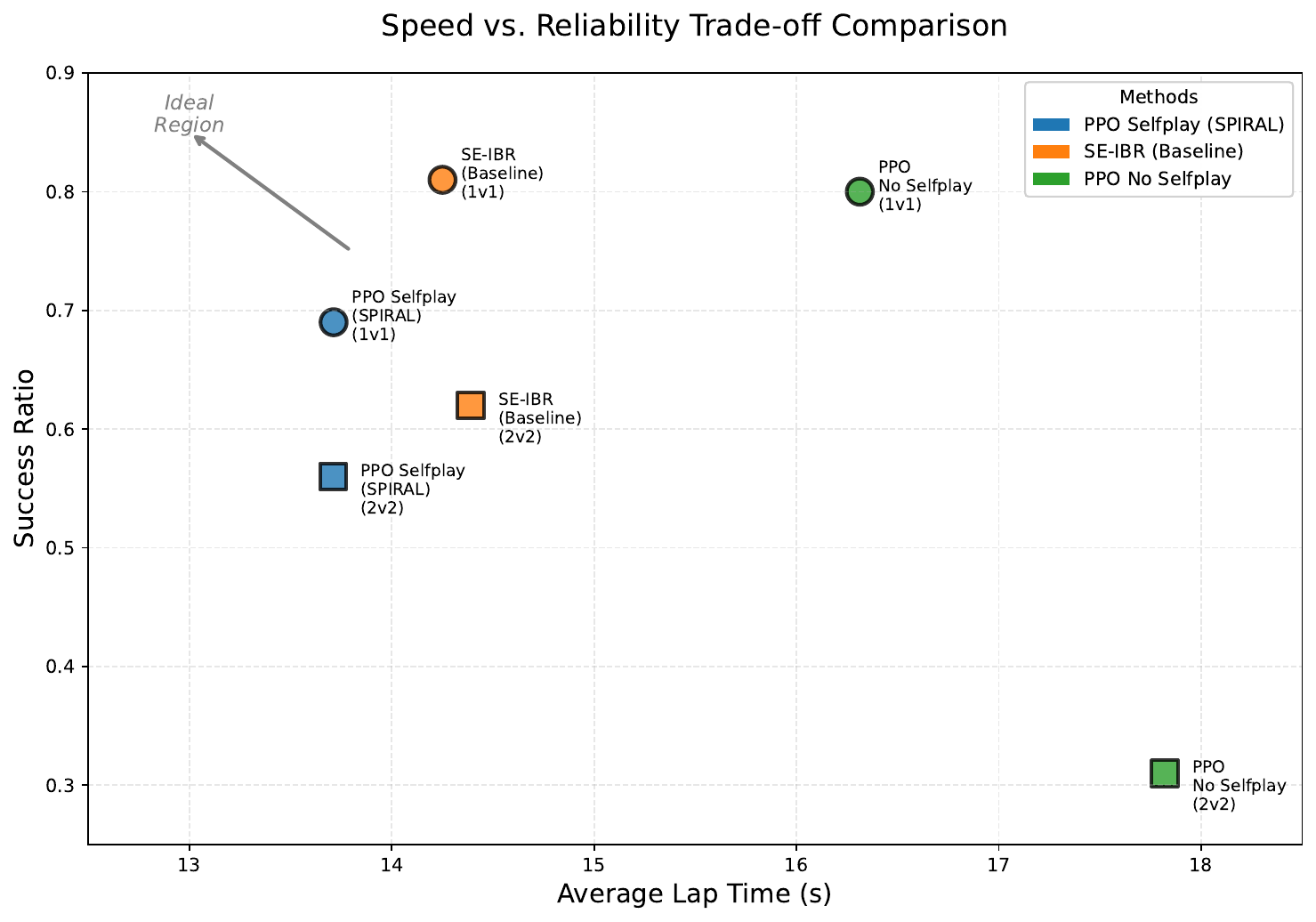}
    \caption{Speed vs. Reliability Trade-off. This plot visualizes the performance of each method in both the 1v1 (circles) and 2v2 (squares) scenarios. The x-axis represents the average lap time (lower is better), and the y-axis represents the success ratio (higher is better). The top-left corner is the "Ideal Region," representing both high speed and high reliability.}
    \label{fig:speed_reliability_tradeoff}
\end{figure}

The results, summarized in Tables \ref{tab:approach_comparison_2_agents_test1} and \ref{tab:approach_comparison_4_agents_test2} and visualized in Figure~\ref{fig:speed_reliability_tradeoff}, reveal a clear and consistent performance trade-off. The plot illustrates how our PPO Self-Play (SPIRAL) agent consistently achieves the fastest lap times, positioning it as the most aggressive and speed-optimized method. In contrast, the more conservative SE-IBR baseline prioritizes course completion, resulting in the highest success ratios at the cost of slower speeds.

This visualization also underscores the critical role of our training paradigm. The standard PPO No Selfplay agent is not only significantly slower but its performance collapses in the more complex four-drone (2v2) scenario. This failure to scale effectively validates that advanced techniques like self-play are essential for developing robust policies capable of handling crowded and competitive environments.

\subsection{Discussion of Results}

The experimental outcomes reveal a distinct and informative trade-off between speed and reliability, with each method occupying a different point on the performance spectrum.

Our proposed method, PPO Self-Play (SPIRAL), consistently achieved superior speed, recording the fastest lap times (approx. 13.71s) with exceptionally low variance. This confirms that its competitive self-play mechanism is highly effective at driving the agent to discover aggressive, speed-optimized policies. However, this pursuit of speed compromised safety, resulting in lower success ratios (dropping from 0.69 to 0.56) as the riskier maneuvers increased the likelihood of collisions.

In direct contrast, the SE-IBR (Baseline) proved to be the most reliable method, achieving the highest success ratios in both experiments (up to 0.81). Its performance profile suggests that its game-theoretic planning foundation produces more conservative and safer navigation strategies, effectively maximizing course completion at the cost of slightly slower lap times.

The performance of the standard PPO without Self-Play highlights the limitations of traditional reinforcement learning in this context. It was the slowest method, and its reliability deteriorated significantly in the second, more challenging experiment (success ratio dropping from 0.80 to 0.31). This failure to generalize validates that robust policies in this domain require either the adaptive challenge provided by self-play or the structured guidance of a planner.

In summary, the results demonstrate a clear performance dichotomy. SPIRAL is unequivocally optimized for maximum speed, making it ideal for competitive scenarios where lap time is the primary metric. Conversely, SE-IBR is optimized for maximum reliability, making it more suitable for applications where successful task completion is paramount.

\label{sec:conclusion_future_work}

\subsection{Conclusion and Future Works}

This paper introduced SPIRAL, a PPO-based training algorithm that leverages self-play to achieve superior speed in autonomous drone racing. Our findings demonstrate a clear performance trade-off: the competitive nature of self-play drives the emergence of aggressive, speed-optimized policies, resulting in the fastest lap times. However, this speed comes at the cost of reliability, leading to lower success ratios compared to the more conservative SE-IBR baseline. This highlights SPIRAL's primary limitations: the aggressive, risk-seeking nature of its learned policies, a decentralized reward structure that inhibits true team cooperation, and a staged curriculum that may limit adaptability to novel conditions. The poor performance of a standard PPO agent validates that advanced paradigms like self-play are essential for this complex domain.

Future research will focus on mitigating these trade-offs to develop more balanced and robust multi-agent systems. Key research directions include:
\begin{enumerate}
\item Enhancing Reliability: Integrating safety constraints or risk-aware objectives to reduce collision rates without sacrificing competitive speed.
\item Fostering Cooperation: Designing sophisticated multi-agent reward structures or explicit communication protocols to enable emergent team-based strategies.
\item Improving Adaptability: Implementing dynamic curricula and online adaptation techniques to improve generalization against unforeseen challenges.
\end{enumerate}
Furthermore, the overarching goal is to create versatile autonomous systems that achieve a synergistic balance of speed, safety, and collaborative intelligence.

\bibliographystyle{IEEEtran} 
\bibliography{references}

\end{document}